\documentclass[10pt,journal,compsoc]{IEEEtran}
 
\usepackage{amsmath}
\usepackage{ntheorem}
\usepackage{algorithm}  
\usepackage{algorithmic}
\usepackage{graphicx}
\usepackage{multirow}
\usepackage{amssymb}
\usepackage{color,xcolor}
\usepackage{ulem}
\usepackage{bbm}
\ifCLASSOPTIONcompsoc
  \usepackage[nocompress]{cite}
\else
  \usepackage{cite}
\fi
\ifCLASSOPTIONcompsoc
 \usepackage[caption=false,font=footnotesize,labelfont=sf,textfont=sf]{subfig}
\else
 \usepackage[caption=false,font=footnotesize]{subfig}
\fi

\newtheorem{theorem}{Theorem}

\newtheorem{definition}{Definition} 

\begin{document}
\title{Online Interaction Detection for Click-Through Rate Prediction}
\author{Qiuqiang~Lin and Chuanhou~Gao,~\IEEEmembership{Senior Member,~IEEE}% <-this % stops a space
\IEEEcompsocitemizethanks{\IEEEcompsocthanksitem Q. Lin and C. Gao are with the School of Mathematical Sciences, Zhejiang University, Hangzhou 310027, China.\protect\\
E-mail: \{11735034, gaochou\}@zju.edu.cn}
% <-this % stops an unwanted space
\thanks{Manuscript received June 25, 2021}}

\IEEEtitleabstractindextext{
\begin{abstract}
    % introduction of CTR prediction
    Click-Through Rate prediction aims to predict the ratio of clicks to impressions of a specific link.
    % difficulties of CTR prediction 
    % 1. sparse and high-dimension inputs
    % 2. feature interactions
    % 3. real-time updates
    This is a challenging task since
    (1) there are usually categorical features, and the inputs will be extremely high-dimensional if one-hot encoding is applied,
    (2) not only the original features but also their interactions are important,
    (3) an effective prediction may rely on different features and interactions in different time periods. 
    % our work
    % 1. an interaction detection algorithm
    % 2. a modification for sequential data
    % 3. a way to make use of the interactions
    To overcome these difficulties, we propose a new interaction detection method, named Online Random Intersection Chains.
    The method, which is based on the idea of frequent itemset mining, detects informative interactions by observing the intersections of randomly chosen samples.
    The discovered interactions enjoy high interpretability as they can be comprehended as logical expressions.
    ORIC can be updated every time new data is collected, without being retrained on historical data.
    What's more, the importance of the historical and latest data can be controlled by a tuning parameter.
    A framework is designed to deal with the streaming interactions, so almost all existing models for CTR prediction can be applied after interaction detection.
    % experimental results
    Empirical results demonstrate the efficiency and effectiveness of ORIC on three benchmark datasets. 
\end{abstract}

}

\maketitle

\IEEEraisesectionheading{\section{Introduction}
\label{sec:introduction}}
% introduction to CTR prediction
\IEEEPARstart{C}{lick}-through rate (CTR) prediction is an important application of machine learning algorithms, 
which aims to predict the ratio of clicks to impressions of a specific link.
It has high commercial value for real-world use such as online advertising and recommendation systems \cite{Graepel10, Covington16, Cheng16, Guo17, Zhou18}.
But it is challenging to make a good prediction of CTR for several reasons.
For example, the input of CTR prediction tasks usually contains a number of categorical features, 
which will be extremely sparse and high-dimensional if one-hot encoding is applied \cite{Shan16, He17}.
And not only the original features but also their interactions are of high importance, 
which makes the problem more difficult \cite{Cheng16, Guo17, Wang17, Lian18}.
What's more, the useful interactions may be time-varied due to the drift of trends or individual interest.

% importance of interactive features
The performance of a CTR prediction model can be significantly improved if appropriate interactions are added into the input \cite{McMahan13, He14, Beutel18, Luo19, Liu20}.
But as the total number of interactions grows rapidly with the number of input features, 
it is difficult or infeasible to take all the interactions into consideration.
Many efforts have been devoted to detect the useful interactions.
For example, some interactions are manually added to the input of Deep Neural Networks \cite{Covington16} and Wide\&Deep \cite{Cheng16}.
This could be laborious and may miss some useful interactions.
Many researchers and practitioners turn to algorithms that can detect interactions automatically.
Several automatic feature interaction methods have been proposed in the last few years, 
such as AutoCross \cite{Luo19}, AutoInt \cite{Song19}, AutoCTR \cite{Song20}, AutoFIS \cite{Liu20} and GLIDER \cite{Tsang20}.

% shortcoming of the existing interaction detection methods
A shortcoming of these methods lies in their high computational cost.
For example, AutoFIS assigns a gate to each feature interaction to control whether it's selected.
So a network taking all the interactions as its input should be trained in the search stage.
AutoCTR has to train and evaluate a large number of networks with different architectures to perform NAS research.
The process of GLIDER may take several hours and significant RAM ($>$150GB) to detect the interactions.
For AutoInt and AutoCTR, another disadvantage is that it's inconvenient to interpret the discovered interactions.
As the interactions are represented by some non-additive functions of the embedding of features inside the model, 
it's hard to interpret which interactions are discovered unless checking the weights layer by layer.

% importance of online algorithms
In practical applications of recommendation systems, new data is collected every day because users continuously interact with the system.
How to make good use of both the latest and historical data is of great significance.
Training a new model only on the latest data is the simplest way, but it's a waste of the historical information.
Fine-tuning the model on hand with the new data is memory- and time-efficient.
However, it ignores the historical data that contains long-term preference signal, thus can easily cause overfitting and forgetting issues \cite{Zhang20}.
Full retraining, which trains the model on both the historical and new data, usually has the best performance but results in a heavy computational burden.
A better way to deal with this problem is online or streaming recommendation, which aims to refresh recommendation models based on real-time user interactions \cite{He16, Subbian16, Chang17}.
A related topic is called sequential or session-based recommendation \cite{Hidasi16, Tang18}, which takes a sequence of items that a user has interacted with as the input, and aims to predict the items that a user will likely interact in the future.
These methods are suitable for sequential data, but may still need retraining when new data comes.

% proposed method: ORIC
Much attention has been paid to interaction detection or online algorithms for recommendation systems.
To the best of our knowledge, however, the topic of online interaction detection is relatively unexplored.
The above-mentioned interaction detection models should be retrained when a new set of data arrives.
Full retraining is computationally expensive, and adding an interaction which was only useful long ago may be harmful to current prediction.
So a practical concern is to perform interaction detection continuously, in which more recent data making a greater contribution than less recent one.
Based on the idea of Random Intersection Chains \cite{lin21}, we propose an approach that can continuously detect the useful interactions without retraining on the historical data.
Random Intersection Chains detect interactions by taking some random samples from the database, then estimating the frequency of patterns in the final intersection by maximum likelihood estimation and calculating their confidence by Bayes' theorem.
In this work, the estimated frequency for a pattern is derived from maximum posterior estimation rather than maximum likelihood estimation.
The historical estimation of frequency could be used as the prior probability, so the method retains the historical knowledge well.
By adjusting the formulation of the prior probability, the relative importance of historical data and the latest data can be controlled.
Also, a framework for integrating the time-varying interactions is designed, with which almost any existing CTR prediction model can be applied after interaction detection.

% main contributions
To summarize, our main contributions in this paper are listed as follows:
\begin{itemize}
  \item We propose an algorithm, named Online Random Intersection Chains, that can detect the meaningful high-order feature interactions explicitly and continuously.
  \item We provide an framework to integrate the discovered interactions into the input of existing CTR prediction models. 
  Thus these models can make use of time-varying interactions.
  \item We analyze the convergence and computational complexity of the proposed interaction detection algorithm. So the effectiveness and efficiency of the algorithm can be guaranteed.
  \item We conduct a series of experiments on three large-scale datasets. 
  The results demonstrate that ORIC is efficient and consistent.
  The found interactions are interpretable and helpful for CTR prediction.
\end{itemize}

% organization of the paper.
The rest of the paper is organized as follows. 
In Section~\ref{sec:preliminaries}, we give the definition of some related concepts, such as CTR prediction, categorical feature interaction and online interaction detection. 
Then we introduce some related work about CTR prediction, interaction detection and online recommendation in Section~\ref{sec:related_work}. 
In Section~\ref{sec:methods}, we formally introduce the algorithm for online interaction detection, named Online Random Intersection Chains, and a framework to make use of the detected interactions in online scenario. 
Some theoretical analyses of the proposed algorithm are then given in Section~\ref{sec:theoretical_analysis}.
In Section~\ref{sec:experiment} we report the results of experiments on three benchmark datasets to verify the efficiency and effectiveness of the algorithm.
Finally Section~\ref{sec:conclusion} concludes this paper.

\section{Preliminaries}
\label{sec:preliminaries}
In this section, we formally give the definition of the related concepts, 
such as CTR prediction, categorical feature interaction and online interaction detection.

% definition of CTR prediction
\begin{definition}
  \textbf{CTR prediction}:
  Given a dataset $D=\{\boldsymbol{X}, \boldsymbol{y}\}$, 
  where $\boldsymbol{X}\in \mathbb{R}^{N\times p}$ contains the features of users and items,
  $\boldsymbol{y}\in \{0, 1\}^N$ indicates the clicks of users to items.
  CTR prediction aims to predict the probability of a user clicking a specific item.
\end{definition}

The $i$-th row of $\boldsymbol{X}$ is denoted by $\boldsymbol{X}_i$ and the $j$-th element of $\boldsymbol{X}_i$ is denoted by $\boldsymbol{X}_{i, j}$.
$\boldsymbol{X}_i$ contains the features of a user and an item.
Each feature is either numerical or categorical, where the categorical features are label-encoded.
So $\boldsymbol{X}_i$ is made up of real numbers and integers.
Usually only a small portion of the recommended items will be clicked by the users.
Therefore, relatively few samples in the dataset are belonging to the positive class, which makes CTR prediction an unbalanced binary classification task.

% definition of categorical feature interaction
\begin{definition}
  \textbf{Categorical feature interaction}:
  If $C_1$, $C_2$, ..., $C_k$ are $k$ categorical features, 
  and $c_1$, $c_2$, ..., $c_k$ are specific categories for corresponding features,
  then ($C_1=c_1$, $C_2=c_2$, ..., $C_k=c_k$) is called a $k$-order categorical feature interaction.
\end{definition}

% Specifically, categorical feature interaction reduces to a binary feature that indicates whether a feature has a specific value if $k$=1.
% And a higher-order categorical feature interaction can be regarded as a cross-product of these binary features,
% which has value 1 if and only if the constituent features are all 1. 
Categorical feature interactions could be viewed in two different ways.
% Firstly, it can be comprehended as a logical expression, which will be TRUE if and only every element in the expression is satisfied.
Firstly, an interaction is a binary feature, which will be assigned 1 if and only every element in the expression is satisfied.
% when its logical expression is TRUE and 0 otherwise.
For example, if an interaction is ($C_1$=0, $C_2$=1), then the interaction is 1 for the sample [$C_1$=0, $C_2$=1, $C_3$=1], but 0 for the sample [$C_1$=0, $C_2$=0, $C_3$=1].
This definition captures the interactions between different features, and help the succeeding model to learn the non-linearity relationships.
One of the advantages of this definition lies in its interpretability.
As an interaction can be regarded as a logical expression, its practical meaning is quite obvious.
This definition coincides the term ``interaction'' used in \cite{Shah14},
and will reduce to the latter if all the input categorical features are binary.
The interaction defined here is also a non-additive interaction \cite{Friedman08, Sorokina08, Song19, Tsang18}, since it can not be represented by a linear combination of lower-order interactions.

Another way is to treat an interaction as a set of items, in which every constituent ($C_i=c_i$) component is regarded as an item.
For instance, the interaction ($C_1$=0, $C_2$=1) is an itemset consisting of two items, and this interaction is contained in the sample [$C_1$=0, $C_2$=1, $C_3$=1], but not in the sample [$C_1$=0, $C_2$=0, $C_3$=1].
From this viewpoint, association rule mining methods can be adopted to detect the interactions \cite{lin21tkde}.

In this paper, we use the terms ``interaction'', ``itemset'' and ``pattern'' interchangeably.

% definition of online CTR prediction
\begin{definition}
  \textbf{Online interaction detection}:
  For $T\ge 1$, let $\boldsymbol{D}_{T}$ be the data collected at time $T$, $\mathcal{I}_{T}$ be the interaction detection model fitted on \{$\boldsymbol{D}_{t}:t\le T$\}.
  Online interaction detection aims to find a map $f$ that $\mathcal{I}_{T}=f(\mathcal{I}_{T-1}, \boldsymbol{D}_{T})$ for $T\ge 2$. 
\end{definition}

Because essential interactions may change as time goes by, interaction detection model should be refreshed with new data continuously.
Intuitively, more recent data has a higher value than less recent one.
But it's also unwise to ignore the information in the historical data.
So it's necessary to have an approach that can detect interactions from both latest and historical data, 
while the relative importance of new data and the histories can be controlled.

In the rest of this paper, subscript ``$t$'' or ``$T$'' is used to identify the time period.
Subscript ``$s$'' means the symbol is corresponding to the pattern $s$, and superscript ``$(c)$'' stands for the class label $c$.
For example, we use $p_{s,t}^{(c)}$ to represent the frequency of pattern $s$ among the samples labeled with $c$ in the data collected at time $t$.

\section{Related Work}
\label{sec:related_work}
\subsection{CTR Prediction}
% CTR prediction models: FM, DNN, Wide & Deep, DeepFM, xDeepFM
The CTR prediction problem could be regarded as a special binary classification task, where the data is highly unbalanced.
Various algorithms have been developed to deal with this problem.
For instance, Factorization machine (FM) associates each feature with a low-dimensional vector, and models all possible interactions by the inner product of the corresponding vectors \cite{Rendle10}.
Field-aware Factorization Machine (FFM) allows each feature to associate with different vectors when interacting with features from different fields \cite{Juan16, Juan17}.
There are also other factorization models, such as Attention Factorization Machine (AFM) \cite{Xiao17}, Neural Factorization Machine (NFM) \cite{He17}, Product-Based Neural Networks (PNN) \cite{Qu18} and Field-weighted Factorization Machine (FwFM) \cite{Pan18}.
Recently deep learning models are quite popular.
Deep Neural Networks (DNN) are used for CTR prediction \cite{Zhang16, Covington16}.
Wide\&Deep is proposed to jointly train wide linear models and deep neural networks \cite{Cheng16}.
To make use of the feature interactions, DeepFM use FM as its ``wide'' part rather than a generalized linear model, 
and it has a shared input to its ``wide'' and ``deep'' parts.
Replace the FM with Compressed Interaction Network (CIN) in DeepFM, then xDeepFM is obtained \cite{Lian18}.

\subsection{Interaction Detection}
% automatic feature interaction selection methods: AutoInt, AutoCTR, AutoFIS, GLIDER
Many automatic interaction detection methods for CTR prediction are proposed in the past few years.
AutoInt \cite{Song19} models the feature interactions by a multi-head self-attentive neural network with residual connections.
AutoCTR \cite{Song20} is based on Neural Architecture Search (NAS).
It modularizes interactions as virtual building blocks and wiring them into a space of direct acyclic graphs, then performs evolutionary architecture exploration to select a good architecture.
AutoFIS \cite{Liu20} introduces a gate for each feature interaction to control whether its output should be passed to the next layer, and retrains the model with the essential interactions.
GLIDER \cite{Tsang20} detects interactions by training a lasso-regularized multilayer perceptron (MLP) on a dataset, then identifying the features that have high-magnitude weights to common hidden units.

% related work about interaction detection
It is also a well-established practice among statisticians fitting models on interactions. 
There exist a number of works dealing with interactive feature selection, especially for pairwise interactions. 
Many of them select interactions by hierarchy principle \cite{bien13, hao14, hao18, agrawal19}. 
Some works are free of the hierarchy assumption. 
For instance, Thanei et al. proposed the xyz algorithm, 
where the underlying idea is to transform interaction search into a closest pair problem which can be solved efficiently in subquadratic time \cite{Thanei18}.
Instead of the hierarchy principle, Yu et al. come up with the reluctant principle, which says that one should prefer main effects over interactions given similar prediction performance \cite{yu19}.
Most of the above-mentioned works concentrate on regression task and numerical features.
On the contrary, random intersection trees \cite{Shah14} detect interactions of binary predictors.
It works by detecting the frequent patterns in the positive class based on random intersections, and estimating the patterns' frequency in negative class based on min-wise hashing.
The idea that detecting the patterns frequent in positive class but infrequent in negative class coincides with association rule mining \cite{agrawal93, agrawal98}.
Random intersection chains \cite{lin21} detect frequent patterns by random intersection as well.
But the frequency in both positive class and negative class is estimated by maximum likelihood estimation, so a more careful selection can be conducted.

\subsection{Online Recommendation}
% related work about recommendation on sequential data
In practice, new data is continuously collected, so recommendation systems need to be retrained with this new data periodically. 
An approach to deal with this problem is called online or streaming recommendation, which aims to update recommendations based on real-time user interactions \cite{He16, Subbian16, Chang17, Zhang20}.
For example, \cite{Chang17} proposes a framework termed sRec to provide explicit continuous-time random process models, and a variational Bayesian approach called recursive mean-field approximation to permit online inference.
\cite{He16} proposes an MF method aimed at learning from implicit feedback effectively and develops an incremental update strategy that instantly refreshes model parameters given new incoming data.
\cite{Zhang20} designs a neural network-based transfer component, which transforms the old model to a new model that is tailored for future recommendations.
Rather than CTR prediction, most of these models aim to predict the rating of a user on an item, and it's the rank that finally matters.
Usually the prediction is mainly based on users' history activities, with few or even no additional features about users or items.

A related topic is called sequential or session-based recommendation \cite{Hidasi16, Tang18}, which takes a sequence of items that a user has interacted with as the input, and aims to predict the items that a user will likely interact in the future.
\cite{Hidasi16} comes up with an RNN for each user’s interaction sequence to capture the interest evolution, and \cite{Tang18} uses CNN as a solution to address the sequential patterns.
These methods are suitable for sequential data, but may still need retraining when new data comes.

\section{Methods}
\label{sec:methods}
\subsection{Online Random Intersection Chains}
% chain generation
A random intersection chain is a linked list.
The head node contains the items in a random sample,
while other nodes contains the intersection of the itemset in its previous node and another random sample.
% data structure
The most intuitional way to store a random intersection chain is recording the itemset for each node.
But it's worth noting that the chain is nonincreasing. 
In other words, the itemset in a node is a subset of the previous node (except the head node).
Thus storing all the itemsets explicitly is wasteful.
To handle this problem, we come up with a special representation for chains.
We use two vectors to represent a chain, denoted by [$item$, $count$].
The first vector $item$ records the items in the head node and the other vector $count$ records how many times the corresponding item occurs in the chain.
In most cases, the input features as well as their orders keep unchanged.
So $item$ can be further simplified as a copy of the first sample, and $count$ is a vector consisting of integers.

\begin{figure}[!t]
  \centering
  \includegraphics[width=\linewidth]{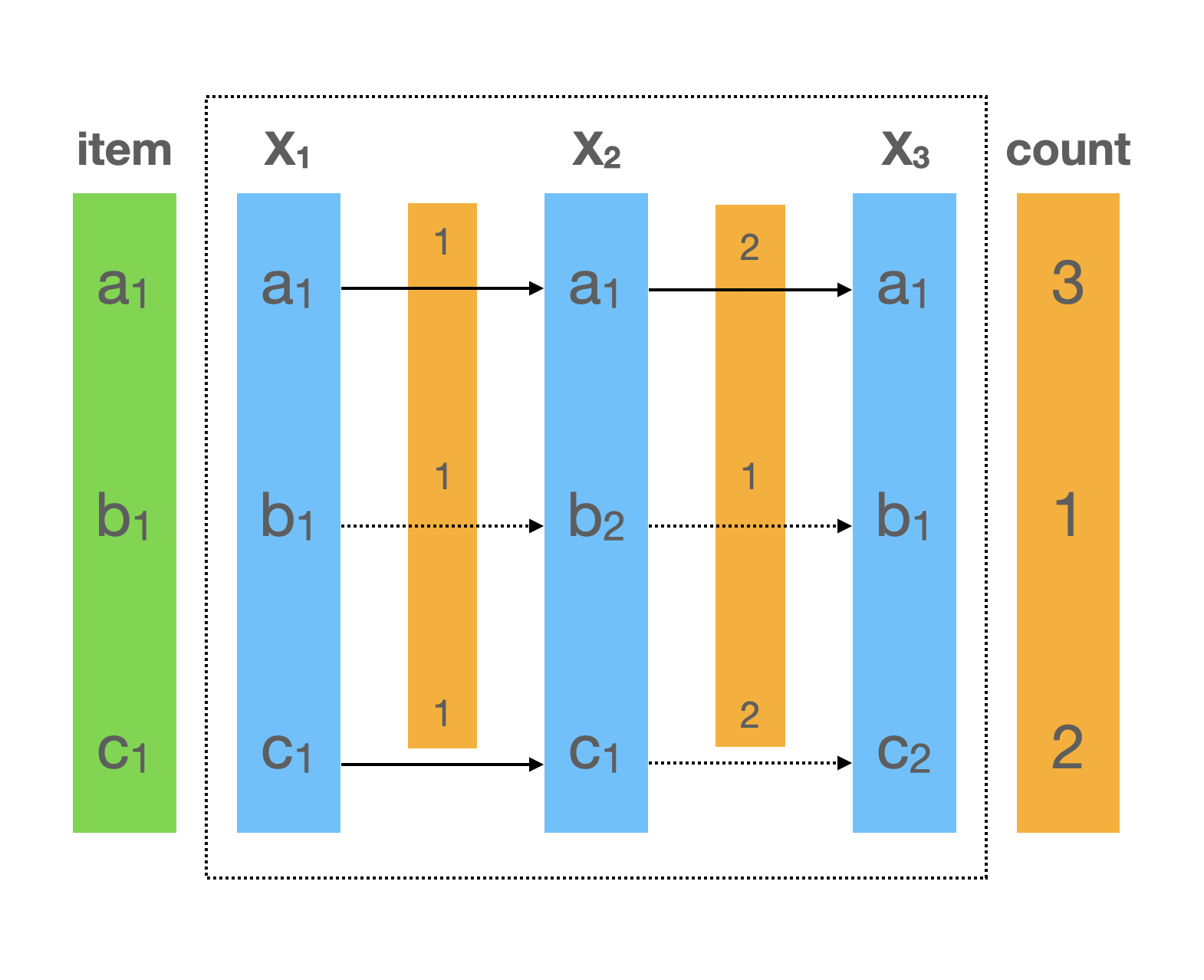}
  \caption{An illustration of chain generation.}
  \label{fig:chain_generation}
\end{figure}

The detail of chain generation can be summarized as follows.
First randomly sample an instance $\boldsymbol{X}_1$, 
set $item$=$\boldsymbol{X}_{i_1}$, and $count$=$\mathbf{1}_p$, where $\mathbf{1}_p$ is an all-ones vector of dimension $p$.
After choosing the $k$-th sample $\boldsymbol{X}_{i_k}$, 
if $count_j$=($k$-1) and $item_j$=$\boldsymbol{X}_{{i_k}, j}$, then set $count_j$=$k$.
These operations will be repeated until the maximum length of a chain is reached or the number of items in the tail node is sufficiently small.
A typical process for chain generation is illustrated in Figure~\ref{fig:chain_generation}.
In this example, three randomly chosen samples are ($A$=$a_1$, $B$=$b_1$, $C$=$c_1$), 
($A$=$a_1$, $B$=$b_2$, $C$=$c_1$) and ($A$=$a_1$, $B$=$b_1$, $C$=$c_2$).
The generated intersection chain is ($A$=$a_1$, $B$=$b_1$, $C$=$c_1$)$\to$($A$=$a_1$, $C$=$c_1$)$\to$($A$=$a_1$), 
which could be represented by [($A$=$a_1$, $B$=$b_1$, $C$=$c_1$), (3, 1, 2)] 
and further simplified as [($a_1$, $b_1$, $c_1$), (3, 1, 2)].

% Similar to random intersection trees, 
% random intersection chains are generated by taking intersection of randomly chosen samples.
% But random intersection trees are originally designed for binary features, 
% and record the itemset for every node separately.
% Since the categorical features for CTR prediction usually have high cardinality, 
% converting them to binary ones is not appealing.
% And it requires a large amount of memory to store the trees when they are deep.
% On the contrary, random intersection chains can get over these difficulties.

After the generation of chains, the itemsets in the tail nodes will be treated as frequent patterns.
% Random intersection trees handle this problem by merely treating the complete interactions in the leaf nodes as frequent patterns, 
% but neglecting their subsets.
% And their frequency is estimated by min-wise hashing other than random intersection trees, 
% to remove the interactions that are also frequent in negative class.
% In fact, every subset of a frequent itemset is frequent, and is possibly confident.
% So it's not an ideal idea to ignore them.
% And estimating frequency by another algorithm may be unnatural and wasteful for ignoring the information in the trees.
% calculation of frequency
% For random intersection chains, however, it's possible to estimate the frequency of every interaction by themselves.
% So a more efficient and precise selection can be executed based on these estimations.
For a pattern with frequency $p$, the distribution of its number of occurrence $k$ in a chain of length $L$ is
\begin{equation}
  \mathbb{P}(k|p)=\begin{cases}
                    p^{k}(1-p), \hfill \mbox{if~} k<L\\
                    p^{k}, \hfill \mbox{if~} k=L
                  \end{cases}.
\end{equation}
And for $M$ such chains, the distribution is
\begin{equation}
  \mathbb{P}(k_1, k_2, ..., k_M|p)=\mathbb{P}(K, I|p)=p^K(1-p)^I,
\end{equation}
where $k_m$ is the number of this pattern's occurrence in the $m$-th chain, 
$K=\sum_{m=1}^Mk_m$ is the number of occurrence of this pattern in total $M$ chains, 
$I=\sum_{m=1}^M\chi_{[k_m<L]}$ is the number of chains that don't contain this pattern in its tail node.

% base version: maximum likelihood estimation
Thus for $M$ chains, the log of likelihood is
\begin{equation}
  \label{eq:loglikelihood}
  \log \mathbb{P}(K, I|p)=K\log p+I\log (1-p).
\end{equation}
Setting the derivative equalling to zero and rearranging, the maximum likelihood estimation of frequency is
\begin{equation}
  \label{eq:frequency_likelihood}
  \hat{p}=\frac{K}{K+I}.
\end{equation}

% online version: maximum posterior
If we have some prior knowledge about the frequency beforehand, e.g. the historical records of a pattern's frequency,
then the frequency could be estimated by maximum posterior estimation.
Assume the prior distribution of a pattern's frequency is subject to a beta distribution, which means
\begin{equation}
  {\rm Beta}(p|a, b)=\frac{\Gamma (a+b)}{\Gamma (a)\Gamma (b)}p^{a-1}(1-p)^{b-1},
\end{equation}
where $\Gamma(\cdot)$ is the Gamma function, $a$ and $b$ are two parameters.
Then the posterior distribution has the form
\begin{equation}
  \mathbb{P}(p|K, I)\propto \mathbb{P}(K, I|p){\rm Beta}(p) \propto p^{K+a-1}(1-p)^{I+b-1}.
\end{equation}
So the posterior distribution $\mathbb{P}(p|K, I)={\rm Beta}(p|K+a, I+b)$ has the same functional form as the prior.
This means beta distribution is indeed a conjugate prior,
thus the calculations can be greatly simplified.
Similar to maximum likelihood estimation, the maximum posterior estimation of frequency is
\begin{equation}
  \label{eq:frequency_posterior}
  \hat{p}=\frac{K+a-1}{K+a+I+b-2}.
\end{equation}

When $T$=1, we have no prior knowledge about the prior.
The parameters $a$ and $b$ are simply set as 1 and the beta distribution reduces to a uniform distribution between 0 and 1.
The posterior distribution in this case is actually equivalent to maximum likelihood estimation.
For $T\ge 2$, the posterior distribution at time $T$-1 can be used as the prior for time $T$.
So the posterior distribution at time $T\ge 2$ is ${\rm Beta}(p|\sum_{t=1}^{T}K_{t}+1, \sum_{t=1}^{T}I_{t}+1)$, where $K_{t}$ and $I_{t}$ are the corresponding statistics for chains generated at time $t$.
This is the same as if all the generated chains are used for the current estimation.
What's more, both $K_{t}$ and $I_{t}$ can be weighted by a coefficient $\gamma\in [0, 1]$, as shown in (\ref{eq:weighted_K_I}).
\begin{equation}
  \label{eq:weighted_K_I}
  \begin{aligned}
    \hat{K}_{T}&=K_{T}+\gamma \hat{K}_{T-1}=\sum_{t=1}^{T}{\gamma^{T-t}K_{t}},\\
    \hat{I}_{T}&=I_{T}+\gamma \hat{I}_{T-1}=\sum_{t=1}^{T}{\gamma^{T-t}I_{t}}.
  \end{aligned}
\end{equation}

The corresponding maximum posterior estimation at time $T$ is (\ref{eq:frequency_posterior_weighted})
\begin{equation}
  \label{eq:frequency_posterior_weighted}
  \hat{p}_{T}=\frac{\hat{K}_{T}}{\hat{K}_{T}+\hat{I}_{T}}.
\end{equation}
For $\gamma<1$, the earlier historical records of $K$ and $I$ have less influence on the current estimation.
Therefore the impact of historical data is limited while some information is still acquired from it.

% calculation of confidence: Bayes theorem
Once frequency is estimated, it's not difficult to calculate the confidence according to the Bayes' theorem as
\begin{equation}
  \label{eq:confidence}
  q_s^{(1)}=\mathbb{P}(Y=1|s\subset X)=\frac{p_s^{(1)}p^{(1)}}{p_s^{(0)}p^{(0)}+p_s^{(1)}p^{(1)}},
\end{equation}
where $p_s^{(c)}$ is the proportion of samples containing pattern $s$ among those with label $c$, 
$p^{(c)}$ is the proportion of samples with label $c$.

The procedure of interaction selection is summarized as follows. 
(i) Divide the database according to the label. 
(ii) Generate chains for positive and negative class separately.
(iii) Estimate the frequency of patterns in the tail nodes of positive class by (\ref{eq:frequency_posterior_weighted}).
(iv) Collect the the most frequent patterns as frequent patterns.
(v) Estimate the frequency of frequent patterns in negative class by (\ref{eq:frequency_posterior_weighted}) and calculate their confidence by (\ref{eq:confidence}).
(vi) Select the most confident patterns as useful interactions.

The hyperparameters in ORIC include the maximum lenght of a chain $L$, the number of chains $M$, the number of frequent interactions $d_{\rm freq}$, the number of confident interactions $d_{\rm conf}$, and the time decay parameter $\gamma$.
Denoting the set of ever detected patterns before time $T$ by $S_T$, the learning parameters at time $T$ are $\hat{K}_{s}^{(c)}$ and $\hat{I}_{s}^{(c)}$ for $c\in \{0,1\}$ and $s\in S_T$.
During the initialization of ORIC, the hyperparameters should be set by hand, and learning parameters are all assigned 0.
When new data is collected, ORIC is updated according to Algorithm~\ref{alg:update_oric}.

\begin{algorithm}
  \caption{Update Online Random Intersection Chains}
  \label{alg:update_oric}
  \begin{algorithmic}[1]
  \REQUIRE 
      $D_{T}$(newly collected data); \\
      $\gamma$(time decay parameter);\\
      \{$\hat{K}_{s,T-1}^{(c)}, \hat{I}_{s,T-1}^{(c)}$\}(current learning parameters)
  \ENSURE 
      \{$\hat{K}_{s,T}^{(c)}, \hat{I}_{s,T}^{(c)}$\}(updated learning parameters)
  \FORALL{$s\in S_T$}
      \FOR{c in $\{0, 1\}$}
          \STATE {$\hat{K}_{s,T-1}^{(c)}\leftarrow \gamma \hat{K}_{s,T-1}^{(c)}$}
          \STATE {$\hat{I}_{s,T-1}^{(c)}\leftarrow \gamma \hat{I}_{s,T-1}^{(c)}$}
      \ENDFOR
  \ENDFOR
  \STATE {Divide $D_{T}$ into $D_{T}^{(0)}$ and $D_{T}^{(1)}$}
  \FOR{c in $\{0, 1\}$}
      \STATE {Generate chains for $D_{T}^{(c)}$}
      % \STATE {$\mathbf{F}^{(c)}\leftarrow$\{frequent patterns for $D_{T}^{(c)}$\}}
      \FORALL{$s$ in tail nodes}
          % \STATE {Count $K_{s,T}^{(c)}$ and $I_{s,T}^{(c)}$ in the chains}
          \STATE {$\hat{K}_{s,T}^{(c)}\leftarrow \hat{K}_{s,T-1}^{(c)}+K_{s,T}^{(c)}$}
          \STATE {$\hat{I}_{s,t}^{(c)}\leftarrow \hat{I}_{s,T-1}^{(c)}+I_{s,T}^{(c)}$}
      \ENDFOR
  \ENDFOR
  % \STATE {Estimate $p_s^{(0)}$ and $p_s^{(1)}$ for $s\in \mathbf{F}^{(1)}$ by (\ref{eq:frequency_posterior_weighted})}
  % \STATE {Estimate $q_s^{(1)}$ for $s\in \mathbf{F}^{(1)}$ by (\ref{eq:confidence})}
  % \STATE {Select the most confident patterns in $\mathbf{F}^{(1)}$}
  \STATE {Return \{$\hat{K}_{s,T}^{(c)}, \hat{I}_{s,T}^{(c)}$\}}
  \end{algorithmic}
\end{algorithm}

% The procedure of interaction selection can be summarized as follows. 
% First the database is divided according to labels, 
% the the number of samples in both positive and negative class is counted. 
% After that, chains are generated for positive and negative class separately.
% Since it's the patterns for positive class that matter, 
% only patterns in the tail nodes of chains for positive class are considered.
% Estimating the frequency of these frequent patterns for both positive class and negative class by Equation~\ref{eq:frequency_likelihood} or Equation~\ref{eq:frequency_posterior}, 
% the confidence can be calculated by Equation~\ref{eq:confidence}.
% Finally the most frequent and confident patterns are used as interactions.

\subsection{Streaming Integrated Model}
Since the proposed interaction detection method is model-agnostic, 
almost all existing models for CTR prediction can be applied after the detection of interactions.
The interactions found by ORIC can be added to the original input directly as binary features.
But this approach has difficulty for online use.
Since the generated interactions evolve with time, training the model with current informative interactions but on the historical data seems redundant and problematic.
On the contrary, if the model is trained only with the latest data, 
it doesn't make full use the historical information.

\begin{figure}[!t]
  \centering
  \includegraphics[width=\linewidth]{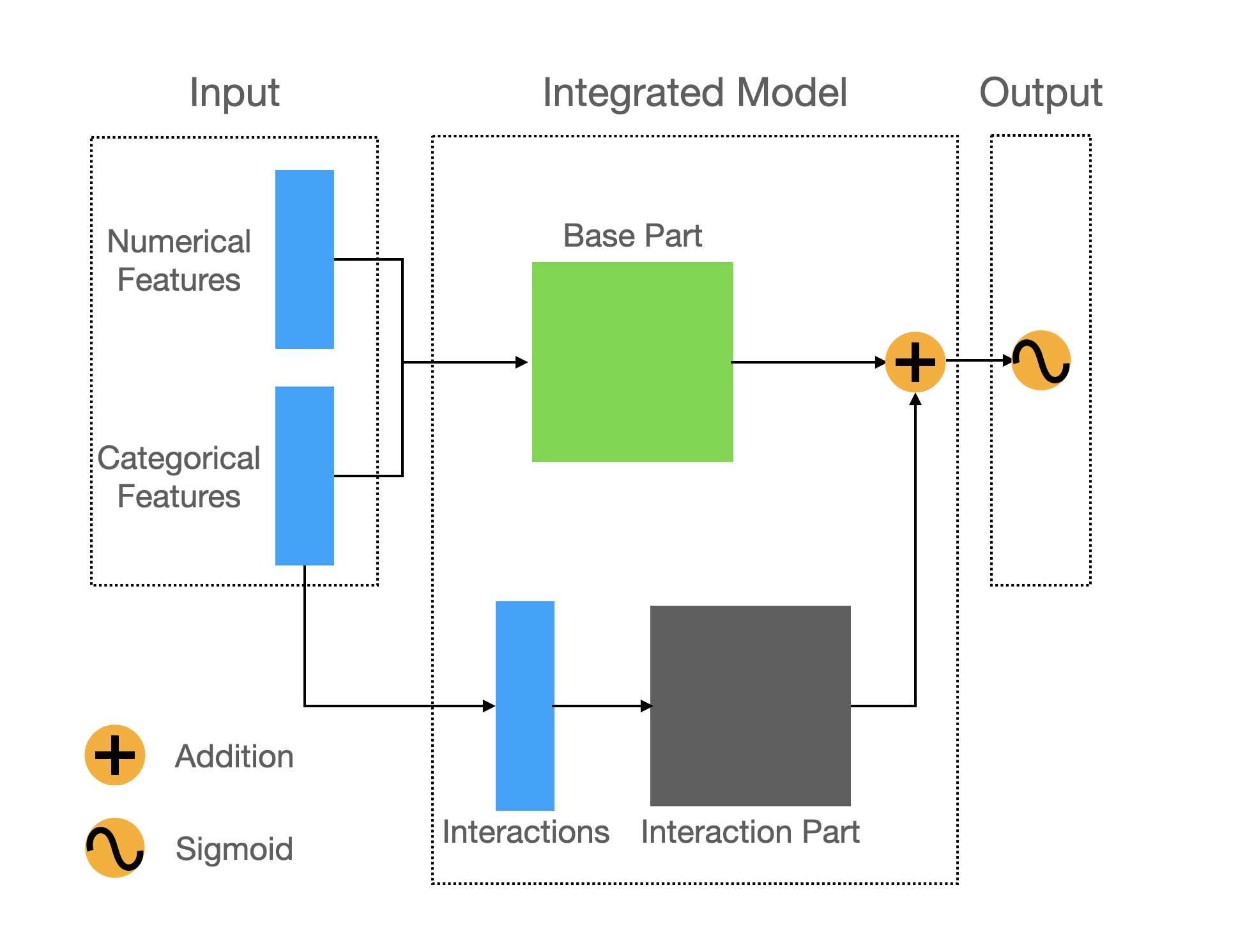}
  \caption{Architecture of the streaming integrated model}
  \label{fig:integrated_model}
\end{figure}

% base part and interaction part: like wide & deep
Inspired by Wide\&Deep learning, we design a generic model framework as illustrated in Figure~\ref{fig:integrated_model}.
This model consists of a base part as well as an interaction part.
The base part takes the original features as input, while the interaction part takes the interactions as input.
Just like the setting in Wide\&Deep, the base part and interaction part are combined using a weighted sum of their output log odds as the prediction.
Both the base part and the interaction part can be any existing CTR prediction model, such as Wide\&Deep, DeepFM, xDeepFM, Deep\&Cross or AutoInt.

Because the input interactions change over time, the interaction part should be initialized every time new data is collected.
However, the input features for base part stay the same. 
To make better use of historical information, we design a workflow analogous to transfer learning as following:
(1) Copy the weights from the previous streaming integrated model or an independent base model to the base part.
(2) Freeze the base part. 
(3) Randomly initialize the weights of interaction part. 
(4) Train the integrated model on the latest data. 
An optional step is fine-tuning, which consists of unfreezing the base part and re-training the whole model on the new data with a low learning rate.

\subsection{Algorithm Overview}
We now provide the general procedure of applying Online Random Intersection Chains and building the streaming integrated model in practical online situations.
Algorithm~\ref{alg:evaluation_update} shows how we evaluate and update the model when new data comes at time $T$. 
First the interactions for newly collected data is generated according to the current interaction detection model.
Then the current integrated model can be evaluated with these interactions.
Next the base model and interaction detection model is updated on the newly collected data.
Finally a new integrated model is built and trained on on the latest data, which will be evaluated when new data is available in the future.

\begin{algorithm}
  \caption{Model evaluation and update}
  \label{alg:evaluation_update}
  \begin{algorithmic}[1]
  \REQUIRE 
      $D_{T}$(newly collected data); \\
      $M_{T-1}$(current base model); \\
      $\tilde{M}_{T-1}$(current integrated model); \\
      $\mathcal{I}_{T-1}$(current interaction detection model)
  \ENSURE 
      $M_{T}$(updated base model);\\
      $\tilde{M}_{T}$(updated integrated model);\\
      $\mathcal{I}_{T}$(updated interaction detection model)
  \STATE {Generate interactions $I_{test,T}$ for $D_{T}$ according to $\mathcal{I}_{T-1}$}
  \STATE {Evaluate $\tilde{M}_{T-1}$ with $D_{T}$ and $I_{test,T}$}
  % \STATE {Divide $D_{,t}$ into $D_{t, train}$ and $D_{t, valid}$}
  \STATE {Fine-tune $M_{T-1}$ with $D_{T}$ to be $M_{T}$}
  \STATE {Update $\mathcal{I}_{T-1}$ with $D_{T}$ to be $\mathcal{I}_{T}$}
  \STATE {Generate interactions $I_{train,T}$ for $D_{T}$ according to $\mathcal{I}_{T}$}
  \STATE {Initialize the integrated model $\tilde{M}_{T}$ with $M_{T}$ as its base part. 
      Then fine-tune $\tilde{M}_{T}$ with $D_{T}$ and $I_{train,T}$}
  \STATE{Return $M_{T}$, $\tilde{M}_{T}$ and $\mathcal{I}_{T}$}
  \end{algorithmic}
\end{algorithm}

\section{Theoretical Analysis}
\label{sec:theoretical_analysis}
In this section, we theoretically analyze some properties of ORIC, such as convergence, the existance of appropriate hyperparameters and its computational complexity. 
Most of the analyses are analogous to those in \cite{lin21} and the proofs are relegated to the Appendix..

One may question the rationality of (\ref{eq:frequency_posterior_weighted}) because $\hat{p}_{s,T}$ seems to be heuristic and in lack of practical meaning at a first glance.
However, Theorem~\ref{thm:freq} shows that $\hat{p}_{s,T}$ is an estimator of an adjusted frequency $\tilde{p}_{s,T}$, which is a weighted average of all historical frequency.
The weight of historical frequency depends on the time stamp and the quantity of itself.
If $0<\gamma<1$, earlier records contributes less to the adjusted frequency. 
And the influence of larger frequency last longer than smaller one.

\begin{theorem}
  \label{thm:freq}
  \it $\hat{p}_{s,T}$ calculated by (\ref{eq:frequency_posterior_weighted}) satisfies:
  \begin{equation}
    \sqrt{M}[\hat{p}_{s,T}-\tilde{p}_{s,T}]\stackrel{d}{\longrightarrow}n(0, \tau^2),
  \end{equation}
  where 
  \[\tilde{p}_{s,T}=\frac{1}{\sum_{t=0}^{T}\alpha_t} \sum_{t=0}^{T}\alpha_tp_{s,t},\]
  \[\alpha_t=\frac{1-p_{s,t}^L}{1-p_{s,t}}\gamma^{T-t},\]
  $\tau^2$ is a positive number depending on $\gamma$ and $p_{s,t}(1\le t\le T)$.
  \hfill
\end{theorem}

Another concern is that whether the frequent patterns can be detected by ORIC.
Due to the randomness of sampling, the algorithm is heuristic.
But according to Theorem~\ref{thm:exist_M_L}, it is guaranteed that the frequent patterns can be detected with arbitrarily high probability, as long as the hyperparameters are appropriately set.

\begin{theorem}
  \label{thm:exist_M_L}
  \it Given $\eta_1, \eta_2 \in (0,1]$, for any $\theta \in (0,1]$, there exist choices of the number of chains $M$, the length of a chain $L$ such that the set of ever detected patterns $S_T$ contains a pattern $s$ with probability at least $1-\eta_1$ if $\tilde{p}_{s,T}^{(c)}\ge \theta$, and with probability at most $\eta_2T$ if $p_{s,t}< \theta$ for all $1\le t\le T$.
  \hfill
\end{theorem}

Because infrequent patterns in positive class will not be selected, keeping an eye on such patterns are useless.
As can be seen from Theorem~\ref{thm:exist_M_L}, there is a small chance for infrequent patterns to be detected by ORIC.
Together with the fact that only two integers (namely $K_{s}$ and $I_{s}$) are stored for pattern $s$, Theorem~\ref{thm:exist_M_L} ensures the little storage space of ORIC.
During the update phase, additional space for $M$ chains is required.
Since a chain is represented by two vectors, the space complexity of an update is $O(M)$ and independent of the length $L$.
According to Theorem~\ref{thm:freq}, the estimation will be more precise for larger $M$.
Thus there is a trade-off between accuracy and efficiency.
If there are many patterns having similar frequency, $M$ should be sufficiently large to obtain accurate frequency.
Contrarily, a few chains are enough when the gap between frequent and infrequent patterns is wide.

\section{Experiments}
\label{sec:experiment}
In this section, experiments are conducted to answer the following questions:\\
(1) Is ORIC efficient enough for large-scale data? How much memory and time will it take to select the interactions?\\
(2) Are the estimations accurate and consistent enough to detect the informative interactions?\\
(3) Is integrating these interactions into the input helpful for existing CTR prediction models in online scenario?\\
(4) Are the detected interactions comprehensible? Do their meanings make sense for human beings?\\

\subsection{Experimental Settings}
We conduct experiments on three public real-world datasets, named Avazu, Criteo and Taobao.
The addresses for downloading the datasets and the experimental codes are given in the Appendix.

\textbf{Avazu: }
This dataset contains the records of whether a displayed mobile ad is clicked by a user or not. 
Click-through data of 10 days, ordered chronologically, is provided.
And the total number of samples is above 40 million.
It has 23 features, all of which are categorical.

\textbf{Criteo: }
This is a benchmark dataset for CTR prediction, which consists of a portion of Criteo's traffic over a period of 7 days. 
There are 45 million users’ clicking records on displayed ads, and the rows are chronologically ordered.
It contains 26 categorical features and 13 numerical features. 

\textbf{Taobao: }
This is a dataset of click rate prediction about display Ad, which is displayed on the website of Taobao. 
1140000 users from the website of Taobao are randomly sampled for 8 days of ad display/click logs to form the original sample skeleton. 
There are 27 million records in the dataset.
13 categorical features and 1 numerical feature are used for making a prediction.

Some important statistics for these datasets are summarized in Table~\ref{tab:statistics}.
\begin{table}
  \renewcommand{\arraystretch}{1.3}
  \caption{Statistics for the datasets.}
  \label{tab:statistics}
  \centering
  \begin{tabular}{ccccc}
    \hline
    Dataset & \#Sample & \#DenseFeat & \#SparseFeat & \#Category\\
    \hline
    Avazu & 40,428,967 & 0 & 23 & 1,544,488 \\
    Criteo & 45,840,617 & 13 & 26 & 998,960 \\
    Taobao & 26,557,961 & 1 & 13 & 2,667,994\\
    \hline
  \end{tabular}
\end{table}

\subsubsection{Data Preprocessing}
% data preprocessing
Both Avazu and Criteo are processed in the same way as provided in \cite{Song19}, which is also adopted by \cite{Song20, Tsang20}.
We ignore the infrequent categories whose appearance is less than a threshold, and label them by a single integer ``0'', which stands for ``others''.
The threshold is 5 for Avazu and 10 for Criteo.
And a numerical value $z$ will be transformed to $(\log z)^2$ if $z>2$. 
As for Taobao, we simply label encode the categorical features and standardize the numerical features.

\subsubsection{Baselines}
% baseline and metrics
To form a streaming integrated model, a basic prediction model is required.
We use a popular CTR prediction model, namely DeepFM, as the basic model.
It is worth mentioning that, other CTR prediction models, such as Wide\&Deep or xDeepFM, can also serve as the basic model.
It is also not necessary for the base part and the interaction part belonging to the same kind of model.
But in this paper, both the base part and the interaction part of the streaming integrated model are DeepFM.

We compare the proposed method with two automatic feature interaction methods for CTR prediction, namely AutoInt and DCN.

\textbf{DCN: }
While keeping the benefits of a DNN model, it introduces a novel cross network that is more efficient in learning certain bounded-degree feature interactions \cite{Wang17}.

\textbf{AutoInt: }
It maps the features into a low-dimensional space, and explicitly models the feature interactions in this low-dimensional space by a multi-head self-attentive neural network with residual connections \cite{Song19}.

Most online recommendation models are aimed at top-N recommendation rather than CTR prediction. 
So we do not compare ORIC with existing online recommendation models.
Instead, two different retraining strategies are adopted to show the effectiveness and efficiency of ORIC:

\textbf{Fine-tuning: }
This method fine-tune the CTR prediction model only on the newly collected data.

\textbf{Retraining-with-reservoir: }
This method maintains a reservoir of historical samples.
When new data comes, we fine-tune the CTR prediction model on both the reservoir and the newly collected data.

% \textbf{Full-retraining: }
% This method trains a new CTR prediction model on all past data every time data is newly collected.

\subsubsection{Evaluation Protocols}
To simulate real-world online scenario, we split each of the data into 10 parts $\{D_1, D_2, ..., D_{10}\}$ with equal size based on their temporary information.
The first 5 parts $\{D_1, D_2, ..., D_{5}\}$ are treated as the ``base training set'', which can be used to train an initial model and perform parameter selection.
The last 5 parts $\{D_6, D_7, ..., D_{10}\}$ are used to evaluate the online algorithms, as if one part is newly collected from one new period.

The procedure of hyperparameter selection and model evaluation can be summarized as follows.
At first, a base model is pre-trained on $D_1,...,D_4$.
Then ORIC with different parameters is fitted on $D_1,...,D_4$, after which a streaming integrated model is built and trained on $D_4$. 
The streaming integrated model is then evaluated and updated on $D_5$ according to Algorithm~\ref{alg:evaluation_update}.
The model with the best parameter is preserved, which will be evaluated and updated on $D_6,...,D_{10}$ according to Algorithm~\ref{alg:evaluation_update}.

Two common metrics for CTR prediction are adopted to evaluate the models, namely AUC (Area Under ROC) and logloss (cross-entropy).
It is noticeable that a small increase of AUC or a slight decrease of logloss at 0.001-level is regarded significant for CTR prediction task, according to existing works \cite{Cheng16, Guo17, Wang17, Song19}.

\subsubsection{Implementation Details}
The structure of all the recommender models are the same as reported in \cite{Tsang20}.
And we use Adam \cite{Kingma14} with learning rate of 0.001 to optimize all deep neural network-based models.
We set all embedding sizes to 16, and the batch size is 8192 for all the cases.
The training is regularized by early stopping to prevent over-fitting.

For ORIC, we set the number of chains $M=10000$, and the number of frequent patterns $d_{\rm freq}=100$.
In order to control the order of discovered interactions directly, we use the size of the tail node rather than the length as the stop criterion.
A chain will stop growing if its tail node contains no more than 4 components.
What's more, according to the reluctant interaction selection principle, one should prefer the lower-order component over a higher-order interaction if all else is equal \cite{yu19, lin21tkde}.
We remove the interactions that are less confident than at least one of its constituents from the result.
So the number of finally selected interactions may be smaller than $d_{\rm conf}$.

The number of confident interactions $d_{\rm conf}$ and time decay parameter $\gamma$ are determined by grid search, where the searching range is $[10,20,...,100]$ for $d_{\rm conf}$ and $[0.0,0.1,...,1.0]$ for $\gamma$.
We also introduce another hyperparameter $\lambda$, which stands for the learning rate after unfreezing the base part if $\lambda>0$ and means the base part will not be unfrozen if $\lambda$=0.
The searching range for $\lambda$ is $[0, 10^{-5}, 10^{-4}, 10^{-3}]$.
We first set $\gamma=1.0$ and $\lambda=0$, and test ORIC with different $d_{\rm conf}$ on the ``base training set''.
Fixing $d_{\rm conf}$ with the best value, we select the best $\gamma$.
After $d_{\rm conf}$ and $\gamma$ are assigned with the best values, $\lambda$ is finally determined.
The best hyperparameters we found in this paper are listed in Table~\ref{tab:hyperparameter}.
\begin{table}[!t]
  \renewcommand{\arraystretch}{1.3}
  \caption{Hyperparameters for the datasets.}
  \label{tab:hyperparameter}
  \centering
  \begin{tabular}{cccc}
    \hline
    Dataset & $d_{\rm conf}$ & $\gamma$ & $\lambda$\\
    \hline
    Avazu & 60 & 0.5 & $10^{-3}$ \\
    Criteo & 30 & 0.0 & $10^{-5}$ \\
    Taobao & 50 & 0.4 & 0 \\
    \hline
  \end{tabular}
\end{table}

The data is preprocessed by Intel(R) Xeon(R) Gold 6148 CPU @2.40GHz,
and the deep models are implemented on a single NVIDIA RTX 2080ti GPU card.
We adopt the implementation of CTR models in a public repository named DeepCTR.

\subsection{Efficiency of ORIC}
In this section, we show that ORIC is both time- and memory-efficient by the analysis of its procedure and the experimental results on three datasets.

\begin{figure}[!t]
  \centering
  \subfloat[Update time]{\includegraphics[height=1.7in]{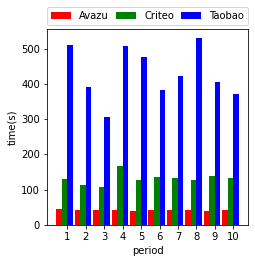}
  \label{fig:time_rico}}
  \hfil
  \subfloat[File size]{\includegraphics[height=1.7in]{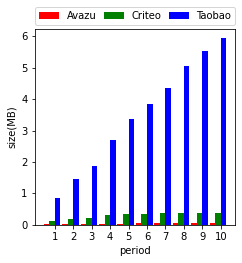}
  \label{fig:size_rico}}
  \caption{Update time and size of ORIC on benchmark datasets.}
  \label{fig:computational_complexity}
\end{figure}

As can be seen from the procedure of chain generation, the running time is mainly affected by the length and total number of chains as well as the number of categorical features.
The number of finally selected interactions $d_{\rm conf}$, time decay parameter $\gamma$ and whether ORIC is previously trained have little influence on the running time.
Once the number of chains $M$ is determined, the running time mainly depends on some statistics of the dataset, e.g. the number of categories and the difference between the frequency of different patterns.
The update time on the 10 parts of each dataset is shown in Figure~\ref{fig:time_rico}.
The average time for updating ORIC is about 40 seconds for Avazu, 2 minutes for Criteo and 8 minutes for Taobao.
We can see that the update of ORIC on Taobao is slowest, although the dataset of Taobao is smallest.
This is because Taobao contains the most categories and many new patterns are detected when new data comes, which requires much time to calculate the frequency.

The demand for memory comes from two sources.
One is to store the parameters of ORIC, the second is to generate the chains.
As analyzed in Section~\ref{sec:theoretical_analysis}, only two integers are stored for a pattern.
Actually the average size of ORIC model for the benchmark datasets is about 40 KB for Avazu, 300 KB for Criteo and 3 MB for Taobao.
The size of ORIC model on each period is shown in Figure~\ref{fig:size_rico}.
Not surprisingly, it is Taobao that consumes the most space due to its large number of categories.
While the size of ORIC on Avazu and Criteo is relatively stable, it increases constantly on Taobao, which indicates that many new patterns are found in every period.
Nevertheless, the space for storing ORIC is almost negligible.
As for chain generation, the chain is represented by two vectors whose dimension is the number of categorical features.
The memory for generating and storing the chains is the same as $2M$ samples.
As stated earlier, $M$ is assigned 10,000 in this paper.
It is as if there is an additional dataset containing 20,000 samples during chain generation, which is very small compared with the original dataset consisting of tens of millions of samples.

\subsection{Consistency of ORIC}
To demonstrate the consistency of ORIC, we exhibit the evolution of the discovered interactions on the validation set $D_5$.
If the detected interactions are exactly the $d_{\rm conf}$ most confident ones, these interactions will be also included in the resulting sets when $d_{\rm conf}$ becomes larger, and the change will be relatively small when $\gamma$ slightly varies.
We use Jaccard-index to evaluate the similarity of interactions found by different values of $d_{\rm conf}$ or $\gamma$. 
The Jaccard-index is defined as 
\begin{equation}
  \label{eq:Jaccard_index}
  {\rm J}(S, S')=\frac{|S\cap S'|}{|S\cup S'|},
\end{equation}
where $S$ and $S'$ are two sets, and $|\cdot|$ stands for cardinality of the set.
The larger Jaccard-index indicates the greater similarity between two sets.

\begin{figure}[!t]
  \centering
  \subfloat[Number of interactions]{\includegraphics[height=1.7in]{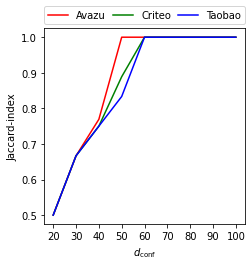}
  \label{fig:jaccard_index_nconf}}
  \hfil
  \subfloat[Time decay]{\includegraphics[height=1.7in]{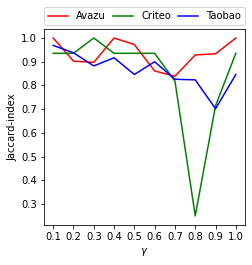}
  \label{fig:jaccard_index_gamma}}
  \caption{Jaccard-indices for adjacent parameters.}
  \label{fig:jaccard_index}
\end{figure}

We first fit ORIC with $\gamma=1.0$ and $d_{\rm conf}$ varying from 10 to 100.
Figure~\ref{fig:jaccard_index_nconf} exhibits the Jaccard-indices of adjacent $d_{\rm conf}$, where the y-axis denotes the Jaccard-index of interactions found with $d_{\rm conf}$ and $(d_{\rm conf}-10)$.
We can see great similarity of detected interactions for close $d_{\rm conf}$. 
The Jaccard-index of a set of cardinality $d_{\rm conf}$ and a set of cardinality $(d_{\rm conf}-10)$ is at most $(d_{\rm conf}-10)/d_{\rm conf}$.
In fact, the Jaccard-index in Figure~\ref{fig:jaccard_index_nconf} is even larger than the upper bound.
This is because among the $d_{\rm conf}$ most confident patterns, some high-order interactions are less confident than its components and dropped in the last.
So the number of finally selected interactions are smaller than $d_{\rm conf}$.
Actually for $d_{\rm conf}\ge 60$, no new interactions are detected even though we enlarge $d_{\rm conf}$.
We check the interactions and ensure that the interactions found with smaller $d_{\rm conf}$ are always contained in the result with larger $d_{\rm conf}$.
Since each ORIC is independently built, these results verify the consistency of ORIC.

Fixing $d_{\rm conf}$ with the best value in Table~\ref{tab:hyperparameter}, Figure~\ref{fig:jaccard_index_gamma} shows the Jaccard-indices between the interactions found by similar $\gamma$, where the y-axis denotes the Jaccard-index of interactions found with $\gamma$ and $(\gamma-0.1)$.
The Jaccard-indices are large in general, which verify the consistency of ORIC again.
There is a sudden fall when $\gamma=0.8$ on Criteo, which means the detected interactions are very different for $\gamma=0.7$ and $\gamma=0.8$.
But for $\gamma=0.9$ and $\gamma=1.0$, large Jaccard-index turns up again, which indicates that patterns found by $\gamma=0.8$, 0.9 and 1.0 are similar.
A possible explanation is that there may be many patterns that are only frequent in one or a few periods.
They will be selected when $\gamma$ is sufficiently large but abandoned otherwise.
Another point is that the interactions vary when the difference of $\gamma$ is large.
For example, the Jaccard-indices of interactions found with $\gamma=0$ and $\gamma=1$ are 0.533, 0.017 and 0.212 on Avazu, Criteo and Taobao, respectively.
This observation indicates the drift of meaningful interactions.

\subsection{Effectiveness of ORIC}
We adopt DeepFM as the base model, while both fine-tuning and retraining-with-reservoir are used as the online updating methods.
Each part of the test data is divided into halves, in which the first half serves as training set and the remaining half is used for validation to prevent overfitting.
For fine-tuning, the models are fine-tuned on the training set.
As for retraining-with-reservoir, we employ the technique of random sampling proposed in \cite{Vitter85}, which is also adopted in SPMF \cite{wang2018}.
A reservoir $R$ of size 2,000,000 is used to maintain some historical data.
The $i$-th instance in the stream is included with probability $|R|/i$, and replaces a random instance in reservoir $R$ if it is included.
After the reservoir is built, we retrain the models on the combination of the training set and the reservoir.
The results of AutoInt and AutoCross, two CTR prediction models that can automatically learn feature interactions, are also given.

Experimental results on the benchmark datasets are shown in Table~\ref{tab:performance}.
As can be seen from the table, the performance of SIM and compared methods verifies that adding interactions found by ORIC and building the integrated model can make a progress.
SIM outperforms the other methods in all cases except Avazu with reservoir.
The comparison between the basic DeepFM model and SIM shows that the improvement exceeds the desired 0.001 level for Criteo and Taobao dataset, which can be regarded as practically significant.

Despite the encouraging results, it should be noted that adding interactions or building an integrated model may cause overfitting.
For Taobao dataset, the loss on training data drops rapidly if we unfreeze the base part and fine-tune the whole SIM, but the loss on validation data or the data in the next period increases.
The test result of fine-tuned SIM is usually worse than it of the basic DeepFM model.
On the contrary, fine-tuning the whole SIM is beneficial for Avazu and Criteo.
Another annoying fact in the online scenario is that the validation set may not represent the future test set well.
Actually for Avazu dataset, the loss on validation set will be smaller if we fine-tune the SIM with a smaller learning rate.
But the test performance is worse even if the validation loss becomes lower.
These observations tell us that although integrating interactions is usually helpful, it should be used with caution.

\begin{table}
  \caption{CTR prediction performance on the benchmark datasets.}
  \label{tab:performance}
  \centering
  \begin{tabular}{ccccc}
    \hline
    Dataset & Method & Model & AUC & logloss\\
    \hline
    \multirow{8}{*}{Avazu} & \multirow{4}{*}{Fine-tune} & DeeoFM & 0.7518 & 0.3918 \\
    & & DCN & 0.7506 & 0.3935 \\
    & & AutoInt & 0.7486 & 0.3933 \\
    & & SIM & \textbf{0.7523} & \textbf{0.3911} \\
    \cline{2-5}
    & \multirow{4}{*}{Reservoir} & DeeoFM & 0.7535 & 0.3902 \\
    & & DCN & 0.7538 & 0.3903 \\
    & & AutoInt & 0.7533 & 0.3902 \\
    & & SIM & 0.7534 & 0.3904 \\
    \hline
    \multirow{8}{*}{Criteo} & \multirow{4}{*}{Fine-tune} & DeeoFM & 0.8019 & 0.4514 \\
    & & DCN & 0.8002 & 0.4525 \\
    & & AutoInt & 0.8029 & 0.4505 \\
    & & SIM & \textbf{0.8046} & \textbf{0.4485} \\
    \cline{2-5}
    & \multirow{4}{*}{Reservoir} & DeeoFM & 0.8009 & 0.4524 \\
    & & DCN & 0.8001 & 0.4528 \\
    & & AutoInt & 0.8020 & 0.4516 \\
    & & SIM & \textbf{0.8046} & \textbf{0.4486} \\
    \hline
    \multirow{8}{*}{Taobao} & \multirow{4}{*}{Fine-tune} & DeeoFM & 0.6024 & 0.2007 \\
    & & DCN & 0.5981 & 0.2010 \\
    & & AutoInt & 0.6005 & 0.2010 \\
    & & SIM & \textbf{0.6093} & \textbf{0.2002} \\
    \cline{2-5}
    & \multirow{4}{*}{Reservoir} & DeeoFM & 0.6088 & 0.2015 \\
    & & DCN & 0.6078 & 0.2014 \\
    & & AutoInt & 0.6073 & 0.2011 \\
    & & SIM & \textbf{0.6116} & \textbf{0.2009} \\
    \hline
  \end{tabular}
\end{table}

\subsection{Interpretability of ORIC}
Due to the comprehensibility of association rules, interactions detected by ORIC are highly interpretable.
Since Avazu contains non-anonymous features, we can understand the meaning of the interactions.
The 10 most confident patterns detected by ORIC in $D_1$ are listed in Table~\ref{tab:detected_interactions}.
To keep notation uncluttered, we only list out the name of features, and omit the specific value for each feature. 
We can see that most of the patterns found by ORIC are pairwise or higher-order interactions, which reflects the fact that there are indeed many essential interactions between different features. 
With a more detailed observation, we can find that many interactions have a feature associated with ``app'' and a feature about ``device'', which indicates the relationship between an item and a user.

\begin{table}
  \caption{Detected interactions for Avazu.}
  \label{tab:detected_interactions}
  \centering
  \begin{tabular}{cc}
    \hline
    RID & Related Features \\
    \hline
    1 & app\_domain, app\_category, device\_conn\_type \\
    2 & app\_category, device\_conn\_type \\
    3 & app\_domain, app\_category \\
    4 & app\_category \\
    5 & app\_id \\
    6 & app\_id, app\_domain \\
    7 & app\_id, device\_conn\_type \\
    8 & app\_id, app\_domain, device\_conn\_type \\
    9 & C1, app\_domain, device\_id \\
    10 & C1, app\_domain, device\_id, device\_type \\
    \hline
  \end{tabular}
\end{table}

For $D_1$, ORIC is not pre-trained and has no historical records.
So the estimated frequency and confidence are supposed to approximate the accurate frequency and confidence in $D_1$.
The estimated frequency in the negative class $\hat{p}^{(0)}$, the estimated frequency in the positive class $\hat{p}^{(1)}$, the estimated confidence $\hat{q}^{(1)}$ are given in Table~\ref{tab:freq_conf}, where the corresponding accurate values are also exhibited.
As can be seen from the table, the estimations are very close to their accurate values, with the numerical order well preserved. 
This observation partially verifies that ORIC can find the most frequent and confident interactions.

\begin{table}
  \caption{Statistics of Detected interactions for Avazu.}
  \label{tab:freq_conf}
  \centering
  \begin{tabular}{ccccccc}
    \hline
    RID & $\hat{p}^{(0)}$ & $p^{(0)}$ & $\hat{p}^{(1)}$ & $p^{(1)}$ & $\hat{q}^{(1)}$ & $q^{(1)}$ \\
    \hline
    1 & 0.5908 & 0.5949 & 0.7112 & 0.7126 & 0.2022 & 0.2014 \\
    2 & 0.5908 & 0.5949 & 0.7112 & 0.7126 & 0.2022 & 0.2014 \\
    3 & 0.6188 & 0.6225 & 0.7417 & 0.7464 & 0.2015 & 0.2016 \\
    4 & 0.6188 & 0.6225 & 0.7417 & 0.7464 & 0.2015 & 0.2016 \\
    5 & 0.6064 & 0.6097 & 0.7243 & 0.7281 & 0.2010 & 0.2009 \\
    6 & 0.6064 & 0.6097 & 0.7243 & 0.7281 & 0.2010 & 0.2009 \\
    7 & 0.5803 & 0.5837 & 0.6933 & 0.6945 & 0.2010 & 0.2003 \\
    8 & 0.5803 & 0.5837 & 0.6933 & 0.6945 & 0.2010 & 0.2003 \\
    9 & 0.5747 & 0.5763 & 0.6757 & 0.6806 & 0.1985 & 0.1992 \\
    10  & 0.5747 & 0.5763 & 0.6757 & 0.6806 & 0.1985 & 0.1992 \\
    \hline
  \end{tabular}
\end{table}

\section{Conclusion}
\label{sec:conclusion}
In this paper, we propose a method to select categorical feature interactions for click-through rate prediction in online scenario.
The common patterns among the positive samples are detected by random intersections, then their frequency is estimated by maximum posterior estimation, where the historical estimation could be used as prior knowledge.
Then their confidence is calculated by Bayes' theorem, and the most confident patterns will be finally selected.
To make full use of the interactions, we construct a streaming integrated model that consists of two parts.
The base part takes the original features as input, while the interaction part is fed with the discovered interactions.
Experimental results on three benchmark datasets demonstrate the effectiveness of the proposed interaction detection methods and the integration approach.

One of the opportunities for future work is adopting more advanced time-series model to predict the future frequency and confidence of an interaction, rather than simply estimating their current values by maximum posterior estimation.
We are also trying to extend ORIC to numerical features.

\ifCLASSOPTIONcompsoc
  \section*{Acknowledgments}
  This work was supported by the National Nature Science Foundation of China under Grant No. 12071428 and 11671418, and the Zhejiang Provincial Natural Science Foundation of China under Grant No. LZ20A010002.
\else
  \section*{Acknowledgment}
\fi

\bibliographystyle{IEEEtran}
\bibliography{IEEEabrv, ms}

\begin{IEEEbiography}[{\includegraphics[width=1in,height=1.25in,clip,keepaspectratio]{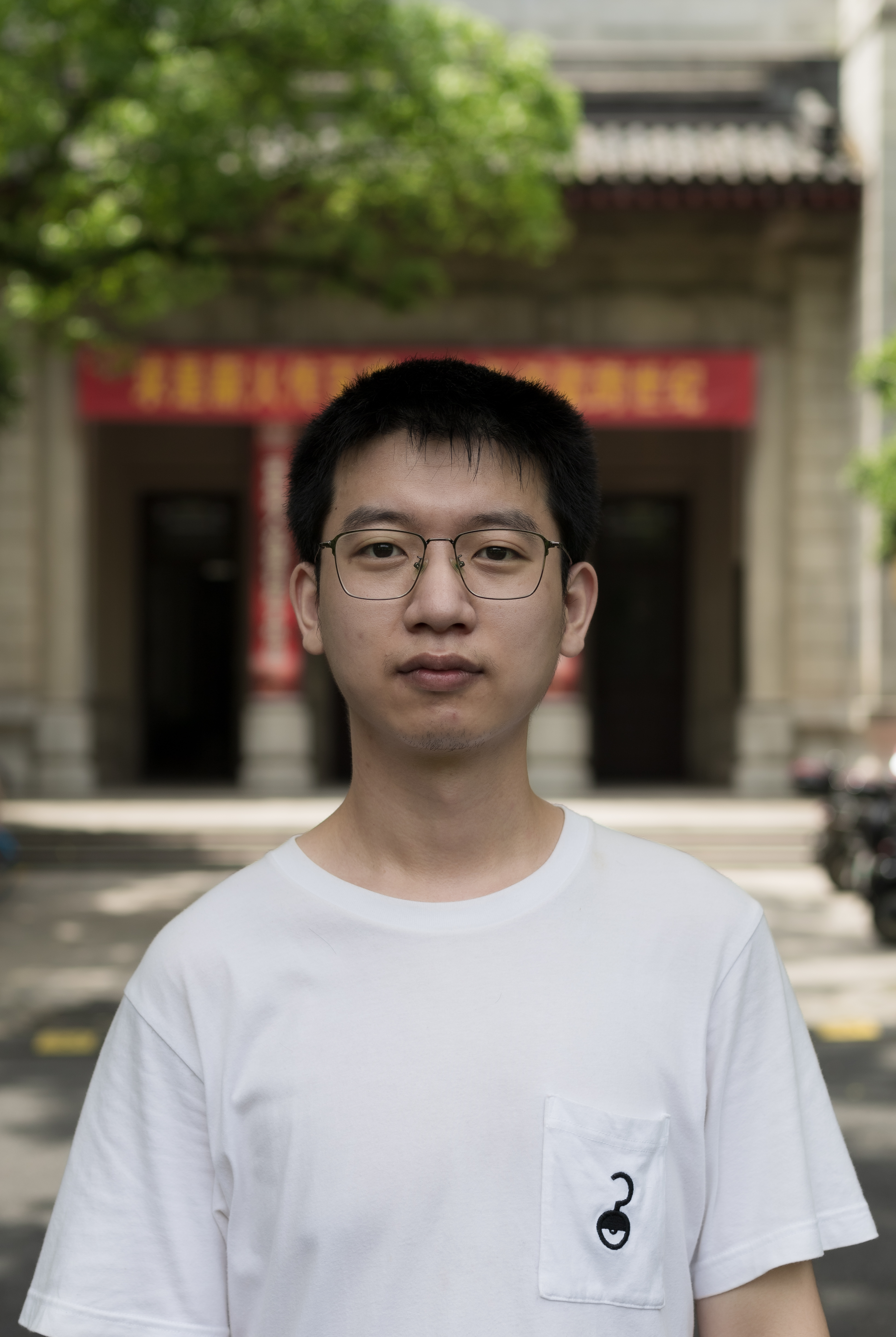}}]{Qiuqiang Lin}
  received the B.S. degrees in Mathematics and Applied Mathematics from Zhejiang University, China, in 2017. He is currently working towards the Ph.D. degree in operational research and cybernetics at Zhejiang University.
  His research interests are in the areas of interaction detection and recommendation systems.
\end{IEEEbiography}

\begin{IEEEbiography}[{\includegraphics[width=1in,height=1.25in,clip,keepaspectratio]{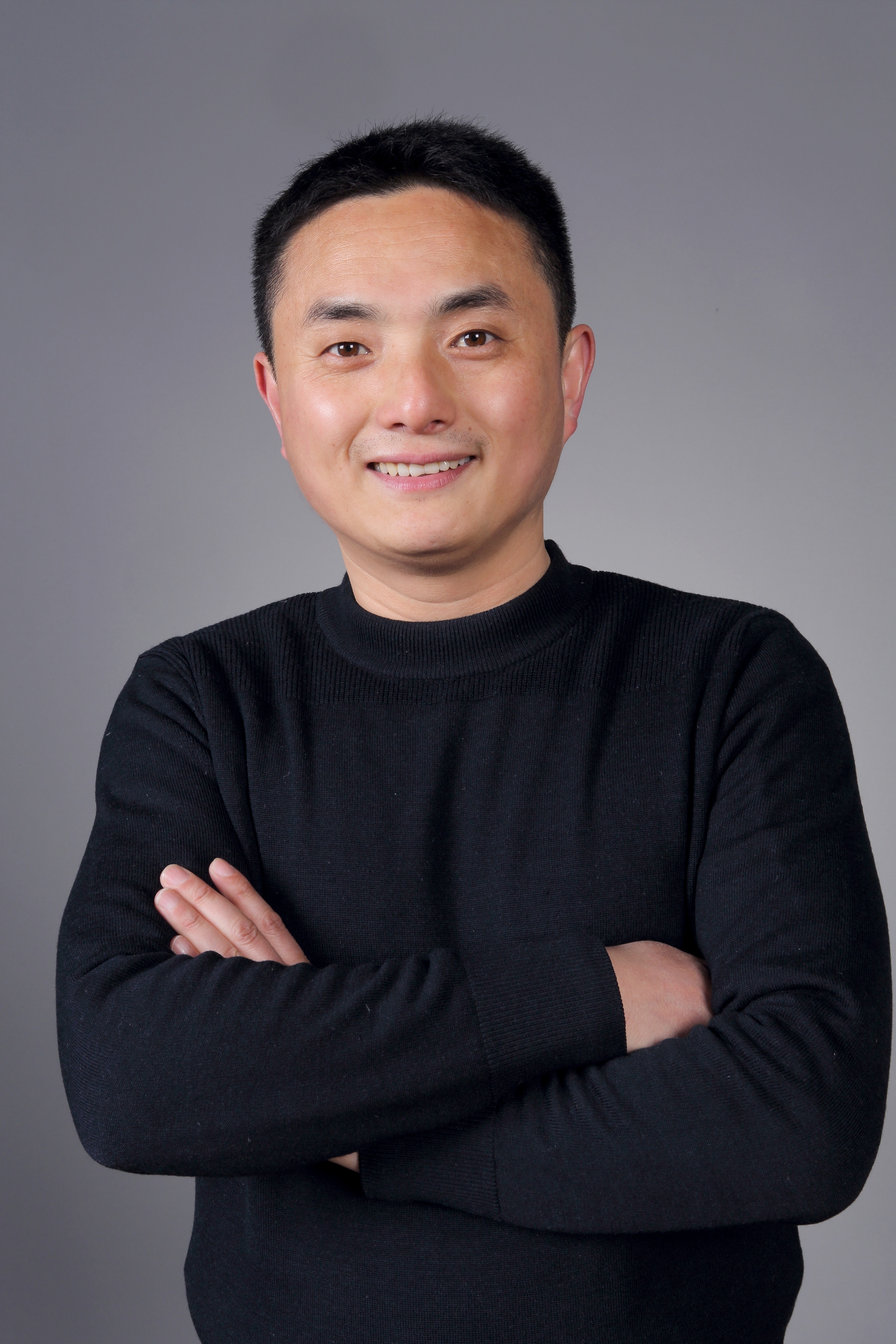}}]{Chuanhou Gao}
  (M’09 SM’12) received the B.Sc. degrees in Chemical Engineering from Zhejiang University of Technology, China, in 1998, and the Ph.D. degrees in Operational Research and Cybernetics from Zhejiang University, China, in 2004. From June 2004 until May 2006, he was a Postdoctor in the Department of Control Science and Engineering at Zhejiang University.\\
  Since June 2006, he has joined the Department of Mathematics at Zhejiang University, where he is currently a Professor. He was a visiting scholar at Carnegie Mellon University from Oct. 2011 to Oct. 2012. His research interests are in the areas of data-driven modeling, control and optimization, chemical reaction network theory and thermodynamic process control. He is an associate editor of IEEE Transactions on Automatic Control from Jan. 2021.
\end{IEEEbiography}

\end{document}

\appendices
\section{Proofs}
In this appendix we give the proofs omitted earlier in the paper.
To keep the notation uncluttered, we omit the subscript ``$s$'' and superscript ``$(c)$'' unless otherwise stated.

\subsection{Proof of Theorem~1}
\begin{IEEEproof}
  For the $m$-th chain generated at time $t$, 
  \begin{equation}
    \mathbb{P}(k_{m,t}|p_t)=\begin{cases}
                      p_t^{k_{m,t}}(1-p_t), \hfill \mbox{if~} k_{m,t}<L\\
                      p_t^{k_{m,t}}, \hfill \mbox{if~} k_{m,t}=L
                    \end{cases}.
  \end{equation}
  For $\chi_{m,t}=\chi_{(k_{m,t}<L)}$, 
  \begin{equation}
    \mathbb{P}(\chi_{m,t}|p_t)=\begin{cases}
                      p_t^L, \hfill \mbox{if~} \chi_{m,t}=0\\
                      1-p_t^{L}, \hfill \mbox{if~} \chi_{m,t}=1
                    \end{cases}.
  \end{equation}
  According to the definition of expectation, we have
  \begin{equation}
    \begin{aligned}
      {\rm E}[k_{m,t}]&=\frac{p_t(1-p_t^L)}{1-p_t}\\
      {\rm E}[\chi_{m,t}]&=1-p_t^{L}.
    \end{aligned}
  \end{equation}
  
  Define 
  \[\hat{k}_{m,T}=\sum_{t=1}^T\gamma^{T-t}k_{m,t},\]
  \[\hat{\chi}_{m,T}=\sum_{t=1}^T\gamma^{T-t}\chi_{m,t},\]
  then we have 
  \begin{equation}
    \begin{aligned}
      {\rm E}[\hat{k}_{m,T}]&=\sum_{t=1}^T\gamma^{T-t}{\rm E}[k_{m,t}]=\sum_{t=1}^T\alpha_tp_t,\\
      {\rm E}[\hat{\chi}_{m,T}]&=\sum_{t=1}^T\gamma^{T-t}{\rm E}[\chi_{m,t}]=\sum_{t=1}^T\alpha_t(1-p_t),
      % {\rm Var}[\chi_{m,t}]&=p_t^{L}(1-p_t^{L})
    \end{aligned}
  \end{equation}

  ${\rm Var}[\hat{k}_{m,T}]$, ${\rm Var}[\hat{\chi}_{m,T}]$ and ${\rm Cov}[\hat{k}_{m,T}, \hat{\chi}_{m,T}]$ could also be calculated by definition, but we omit their actual formulas for simplicity. 
  $k_{m,T}$ can be treated as a random sample for variable $\hat{k}_{T}$, and $\chi_{m,T}$ a random sample for variable $\hat{\chi}_{T}$.
  % Since the chains are generated independently, ${\rm Cov}[k_{m,t}, k_{m',t'}]$ is nonzero only when $m=m'$ and $t=t'$. 

  Define 
  \[g(k,\chi)=\frac{k}{k+\chi},\]
  then we have 
  \[\hat{p}_T=g(\hat{K}_T,\hat{I}_T)=g(M\bar{\hat{k}}_T,M\bar{\hat{\chi}}_T)=g(\bar{\hat{k}}_T,\bar{\hat{\chi}}_T)\]
  \[\tilde{p}_T=g({\rm E}[\hat{k}_T],{\rm E}[\hat{\chi}_T])=\frac{1}{\sum_{t=0}^{T}\alpha_t} \sum_{t=1}^{T}\alpha_tp_{t}\]
  Denote
  \[g_k':=\frac{\partial g}{\partial k}({{\rm E}[k_{m,T}],{\rm E}[\chi_{m,T}}]), \]
  \[g_\chi':= \frac{\partial g}{\partial \chi}({{\rm E}[k_{m,T}],{\rm E}[\chi_{m,T}}]), \]
  \[\tau^2:= (g_k')^2{\rm Var}[\hat{k}_T]+(g_\chi')^2{\rm Var}[\hat{\chi}_T]+2g_k'g_\chi'{\rm Cov}[\hat{k}_T,\hat{\chi}_T], \] 
  then by the Multivariate Delta Method, we have
  \[\sqrt{M}(\hat{p}_T-\tilde{p}_T)\to n(0,\tau^2) \]
  in distribution.
\end{IEEEproof}

\subsection{Proof of Theorem~2}
To prove Theorem 2, we show Theorem 1 in \cite{lin21} here as Lemma \ref{lm:exist_M_L}.
% and leave out the proof, which can be found in \cite{lin21}.
\begin{lemma}
  \label{lm:exist_M_L}
  \it Given $\eta_1, \eta_2 \in (0,1]$, for any $\theta \in (0,1]$, there exist choices of $M$, $L$ such that $s$ appears in at least one of the tail nodes with probability at least $1-\eta_1$ if $P(s\subseteq X|Y=c)\ge \theta$, and with probability at most $\eta_2$ if $P(s\subseteq X|Y=c)< \theta$.
  \hfill
\end{lemma}

\begin{IEEEproof}[Proof of Lemma \ref{lm:exist_M_L}]
% To keep the notation uncluttered, we omit the superscript ``$(c)$'' on the probabilities.
% For a pattern $s$, we use the notation $p_s$ for the probability of $s$'s occurrence conditioned on $Y=c$.
Define $p_1$ as the smallest pattern frequency above $\theta$, $p_2$ as the largest pattern frequency below $\theta$. That is, 
\[ p_1=\min\{p_s:p_s\ge \theta\},\]
\[ p_2=\max\{p_s:p_s<\theta\}.\]
For a pattern $s$, and a chain of length $L$, 
\[\mathbb{P}(s\subseteq S_{L,1})=p_s^L. \]
For a pattern $s$, and $M$ chains of length $L$ , 
\begin{equation}
  \begin{aligned}
    g(p_s;L,M)&=\mathbb{P}(s\subseteq S_{L,M})\\
    &=1-[1-\mathbb{P}(s\subseteq S_{L,1})]^M\\
    &=1-[1-p_s^L]^M.
    \nonumber
  \end{aligned}
\end{equation}
We can see that $g(p_s;L,M)$ is monotone increasing with the increasing of $p_s$ and $M$, and the decreasing of $L$.
For $p_s\ge \theta$, if $M\ge \frac{{\log}\eta_1}{\log(1-p_1^L)}$, then 
\begin{equation}
  \begin{aligned}
    \mathbb{P}(S\subseteq S_{L,M})&=g(p_s;L,M)\\
    &\ge g(p_1;L,\frac{{\log}\eta_1}{\log(1-p_1^L)})\\
    &=1-[1-p_1^L]^{\frac{{\log}\eta_1}{\log(1-p_1^L)}}\\
    &=1-\eta_1.
    \nonumber
  \end{aligned}
\end{equation}
Define 
\[M^*(L)=\lceil \frac{\log\eta_1}{\log(1-p_1^L)}\rceil, \]
\[\bar{M}(L)=\frac{\log\eta_1}{\log(1-p_1^L)}+1. \]
Thus $\bar{M}(L)\ge M^*(L)\ge \frac{\log\eta_1}{\log(1-p_1^L)}$. Then for $p_s\ge \theta$, we have $\mathbb{P}(S\subseteq S_{L,M})\ge 1-\eta_1$ if $M\ge M^*(L)$.

Next we give the conditions for tail nodes containing $s$ with probability at most $\eta_2$ if $P(s\subseteq X|Y=c)< \theta$.
Fixing $M=M^*(L)$, for $p_s< \theta$ we have
\begin{equation}
  \begin{aligned}
    \mathbb{P}(S\subseteq S_{L,M^*})&=g(p_s;L,M^*)\\
    &< g(p_2;L,\bar{M})\\
    &=1-[1-p_2^L]^{\frac{{\log}(\eta_1)}{\log(1-p_1^L)}+1}\\
    &=1-\eta_1^{\frac{{\log}(1-p_2^L)}{{\log}(1-p_1^L)}}(1-p_2^L).
  \end{aligned}
  \nonumber
\end{equation}
Define
\[f(L)=\frac{{\log}(1-p_2^L)}{{\log}(1-p_1^L)}.\]
Take the derivative of $f$, then we have
\begin{equation}
  \begin{aligned}
    f'(L)&= \frac{-p_2^L\log p_2}{1-p_2^L}\log(1-p_1^L)+\frac{p_1^L\log p_1}{1-p_1^L}\log(1-p_2^L)\\
    &= \log(1-p_1^L)\log(1-p_2^L)[f_1(p_1^L)-f_1(p_2^L)],
  \end{aligned}
  \nonumber
\end{equation}
where 
\[f_1(x)= \frac{x\log x}{(1-x)\log(1-x)}.\]
So the corresponding derivative is
\[f_1'(x)= \frac{(1+\log x-x)\log(1-x)+x\log x}{[(1-x)\log(1-x)]^2}.\]
Denote the numerator as $f_2(x)$, and take the derivative, then we have
\[f_2(x)=(1+\log x-x)\log(1-x)+x\log x,\]
\[f_2'(x)= \frac{(1-x)^2\log(1-x)-x^2\log x}{x(1-x)}.\]
Again denoting the numerator as $f_3(x)$ and taking the derivative, we have
\[f_3(x)=(1-x)^2\log(1-x)-x^2\log x,\]
\[f_3'(x)=-2(1-x)\log(1-x)-2x\log x-1.\]
Denoting $f_4(x)=f_3'(x)$, we have
\[f_4'(x)=2\log(1-x)-2\log x=2\log(\frac{1}{x}-1). \]
Therefore for $x\in (0,1)$,
\begin{equation}
  \begin{aligned}
    &f_4'(x)<0~{\rm for}~x\in (0, \frac{1}{2}),~f_4'(x)>0~{\rm for}~x\in (\frac{1}{2}, 1)\\
    \Rightarrow &f_3'(x)= f_4(x)\le f_4(\frac{1}{2})=-1<0\\
    \Rightarrow &f_3(x)\le \lim_{x\to 0}f_3(x)=0\\
    \Rightarrow &f_2'(x)\le 0\\
    \Rightarrow &f_2(x)\le \lim_{x\to 0}f_2(x)=0\\
    \Rightarrow &f_1'(x)\le 0.
  \end{aligned}
  \nonumber
\end{equation}
Noticing that $0\le p_2<p_1\le 1$, we have $f_1(p_1^L)<f_1(p_2^L)$, and thus $f'(L)<0$.
So $g(p_2;L,\bar{M})$ is a monotone decreasing function of $L$.
Extend the domain of $f$ to real numbers, according to $L'H\hat{o}pital's$ rule,
\begin{equation}
  \begin{aligned}
    \lim_{x\to \infty}f(x)&=\lim_{x\to \infty}{(\frac{-p_2^x\log p_2}{1-p_2^x}/\frac{-p_1^x\log p_1}{1-p_1^x})}\\
    &=\frac{\log p_2}{\log p_1} \lim_{x\to \infty}(\frac{p_2}{p_1})^x \lim_{x\to \infty}\frac{1-p_1^x}{1-p_2^x}\\
    &=\frac{\log p_2}{\log p_1}\cdot 0 \cdot 1 = 0.
  \end{aligned}
\end{equation}
Then according to Heine theorem, 
\[\lim_{L\to \infty}f(L)=0.\]
Thus we have
\[\lim_{L\to \infty} g(p_2;L,\bar{M})=1-\lim_{L\to \infty}\eta_1^{f(L)}(1-p_2^L)=0.\]
So for any $\eta_2\in (0,1)$, there exists $L^*\in \mathbb{N}$ such that $\mathbb{P}(S\subseteq S_{L,M^*}^{(c)})\le \eta_2$ if $L\ge L^*$. 
\end{IEEEproof}

\begin{IEEEproof}[Proof of Theorem~2]
  Since $\tilde{p}_{s,T}$ is a weighted average of $(p_{s,1}, ..., p_{s,T})$, $\tilde{p}_{s,T}\ge \theta$ implies that there exists a $t\in \{1,2,...,T\}$ such that $p_{s,t}\ge \theta$.
  According to Lemma \ref{lm:exist_M_L}, $s$ will be detected with probability at least $1-\eta_1$ at time $t$.
  As for a pattern with $p_{s,t}<\theta$ for all $1\le t\le T$, Lemma \ref{lm:exist_M_L} indicates that the probability of not detecting it at one update is larger than $1-\eta_2$. Thus the probability of not detecting it in all $T$ updates is at least $(1-\eta_2)^T$. That is, it will be included in $S_T$ with probability less than $1-(1-\eta_2)^T\le \eta_2T$.
\end{IEEEproof}

\section{Related Links}
\textbf{Avazu dataset:}\\
https://www.kaggle.com/c/avazu-ctr-prediction/data.\\
\textbf{Criteo dataset:}\\
https://www.kaggle.com/c/criteo-display-ad-challenge.\\
\textbf{Taobao dataset:} \\
https://tianchi.aliyun.com/dataset/dataDetail?dataId=56.\\
\textbf{DeepCTR:}\\
https://github.com/shenweichen/DeepCTR.\\
\textbf{Repeatable experiment code:}\\
https://github.com/Lin-John/ORIC

\bibliographystyle{IEEEtran}
\bibliography{IEEEabrv, supplement}